\documentclass[10pt,twocolumn,letterpaper]{article}

\usepackage[pagenumbers]{cvpr} % To force page numbers, e.g. for an arXiv version

\usepackage{graphicx}
\usepackage{amsmath}
\usepackage{amssymb}
\usepackage{booktabs}

\usepackage[pagebackref,breaklinks,colorlinks]{hyperref}

\usepackage[capitalize]{cleveref}
\crefname{section}{Sec.}{Secs.}
\Crefname{section}{Section}{Sections}
\Crefname{table}{Table}{Tables}
\crefname{table}{Tab.}{Tabs.}

\def\cvprPaperID{2484} % *** Enter the CVPR Paper ID here
\def\confName{CVPR}
\def\confYear{2022}

\begin{document}

\title{IIP-Transformer: Intra-Inter-Part Transformer for Skeleton-Based Action Recognition}

\author{Qingtian Wang\textsuperscript{1}, Jianlin Peng\textsuperscript{1,3}, Shuze Shi\textsuperscript{1,2}, Tingxi Liu\textsuperscript{1}, Jiabin He\textsuperscript{1}, Renliang Weng\textsuperscript{1}\\
\textsuperscript{1}Algorithm Research, Aibee Inc\\ \textsuperscript{2}Beijing Jiaotong University\\
\textsuperscript{3}Nanjing University of Aeronautics and Astronautics\\
{\tt\small\{qtwang,jlpeng,tliu,jhe,rlweng\}@aibee.com,19120306@bjtu.edu.cn}
}

%\author{Qingtian Wang\textsuperscript{1}\hspace{5pt}Jianlin %Peng\textsuperscript{1,3}\hspace{5pt}Shuze %Shi\textsuperscript{1,2}\hspace{5pt}Tingxi %Liu\textsuperscript{1}\hspace{5pt}Jiabin %He\textsuperscript{1}\hspace{5pt}Renliang Weng\textsuperscript{1}\\
%\textsuperscript{1}Algorithm Research, Aibee Inc\\ %\textsuperscript{2}Beijing Jiaotong University\\
%\textsuperscript{3}Nanjing University of Aeronautics and %Astronautics\\
%{\tt\small\{qtwang,jlpeng,tliu,jhe,rlweng\}@aibee.com, %19120306@bjtu.edu.cn}
%}

%% omit the following lines up until the closing ``}''.
%% Additional authors and addresses can be added with ``\and'',
%% just like the second author.
%% To save space, use either the email address or home page, %not %both
%\and

\maketitle

%%%%%%%%% ABSTRACT
\begin{abstract}
   Recently, Transformer-based networks have shown great promise on skeleton-based action recognition tasks. The ability to capture global and local dependencies is the key to success while it also brings quadratic computation and memory cost. Another problem is that previous studies mainly focus on the relationships among individual joints, which often suffers from the noisy skeleton joints introduced by the noisy inputs of sensors or inaccurate estimations. To address the above issues, we propose a novel Transformer-based network (IIP-Transformer). Instead of exploiting interactions among individual joints, our IIP-Transformer incorporates body joints and parts interactions simultaneously and thus can capture both joint-level (intra-part) and part-level (inter-part) dependencies efficiently and effectively. From the data aspect, we introduce a part-level skeleton data encoding that significantly reduces the computational complexity and is more robust to joint-level skeleton noise. Besides, a new part-level data augmentation is proposed to improve the performance of the model. On two large-scale datasets, NTU-RGB+D 60 and NTU RGB+D 120, the proposed IIP-Transformer achieves the-state-of-art performance with more than $ 8\times $ less computational complexity than DSTA-Net, which is the SOTA Transformer-based method.
\end{abstract}

%%%%%%%%% BODY TEXT
\section{Introduction}

Human action recognition has been studied during the past decades and achieves promising progress in many applications ranging from human-computer interaction to video retrieval\cite{Carreira_2017_CVPR,2019Gesture,DBLP:conf/iccv/Feichtenhofer0M19,shi2021action}. Recently, skeleton-based representation has received increasing attention due to its compactness of depicting dynamic changes in human body movements~\cite{1973Visual}. Advanced pose estimation algorithms~\cite{2018OpenPose} and advances in the somatosensory cameras such as Kinect~\cite{DBLP:journals/ieeemm/Zhang12} and RealSense~\cite{2017Intel} make it possible to obtain body keypoints accurately and quickly at a low cost. In addition, skeleton-based representation is more robust to variations of illumination and background noises in contrast to RGB representation. These merits attract researchers to develop various methods to exploit skeleton for action recognition.

\begin{figure}[t]
  \centering
  \includegraphics[width=0.8\linewidth]{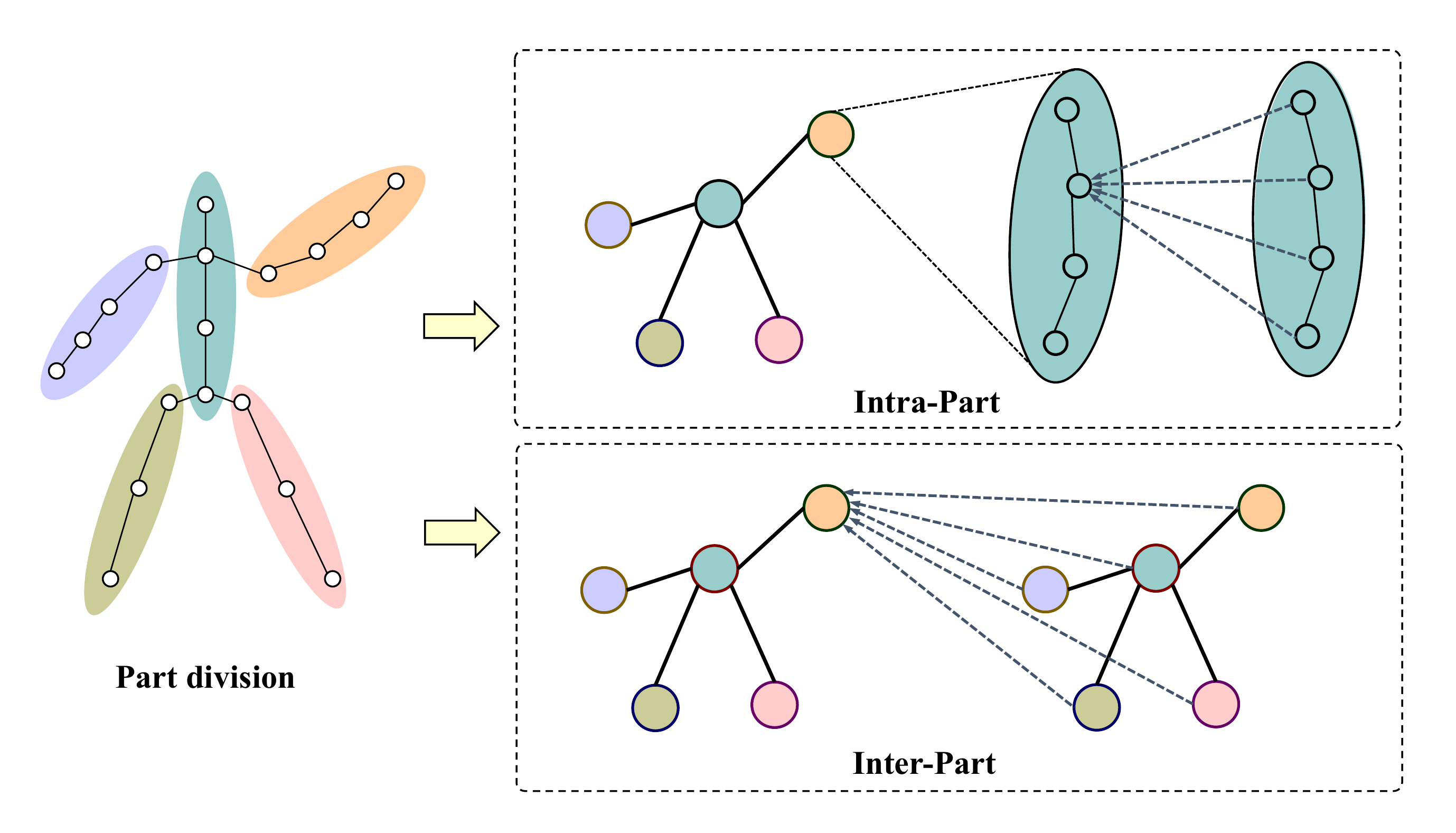}
  \caption{Illustration of our main idea. The body joints are divided into 5 parts. The Inter-Part branch is used to explore relationships between parts and the Intra-Part branch aims to capture dependencies between joints in the same part.}
  \label{fig:1}
\end{figure}

Previous skeleton-based action recognition methods utilize graph topologies or manually designed rules to transform the raw skeleton sequence into a grid-shape structure such as pseudo-image\cite{ding2017investigation, liu2017enhanced, li2018co} or graph, and then feed it into RNNs, CNNs, or GCNs~\cite{DBLP:conf/aaai/YanXL18} to extract features. However, there is no guarantee that the hand-crafted structure is the best choice of capturing joints relationships, which limits the generalizability and performance of previous works. Recently, Transformer-based methods\cite{plizzari2021skeleton, shi2020decoupled} have been proposed, relying on the multi-head self-attention mechanism which adaptively explores the potential dependencies between skeleton joints. Specifically, Shi \etal~\cite{shi2020decoupled} treat each individual joint as a token while the calculation of self-attention grows quadratically to the number of tokens, thus introduces a huge amount of calculations. This work also suffers from the noisy skeleton joints collected by sensors or inaccurate estimations.

To solve these problems, we introduce the concept of body parts into transformer network. Most of actions such as standing up or dancing are performed by co-movement of body parts. Actually, body parts can be considered as the minimum units of action execution, which means these actions can be identified only by the movement of body parts. Different from other complex partition strategies, we simply aggregate body joints into several parts according to human body topologies and encourage the model to exploit the complicated interactions. Specifically, we divide $ v $ body joints into $ p $ parts and encode each part into a token, which reduces the spatial self-attention computation cost by $ v^2/p^2 $ times. Another advantage of our proposed partition encoding is that it enables the model to take sparser frames as temporal inputs, which brings in additional computation reduction. To encourage the model to reason globally instead of relying on a particular part, we propose a new data augmentation method named Part-Mask which masks out a part randomly during training. This new strategy makes the model more robust across challenging cases.

Since the body joints are divided into parts, the joint-level information may be lost. For some fine-grained actions, \eg clapping or writing, it is necessary to capture the interactions between body joints additionally. We propose the novel Intra-Inter-Part Transformer network (IIP-Transformer) to tackle this issue and make three main improvements comparing with standard Transformer networks. First, Intra-Inter-Part self-attention mechanism is proposed to simultaneously capture intra-part features and inter-part features without increasing much computational complexity, as depicted in \Cref{fig:1}. Second, inspired by BERT~\cite{devlin2018bert}, we introduce a learnable class-token instead of pooling all features extracted by backbone. Last, instead of using two individual transformers to model spatial and temporal dependencies, we propose a new spatial-temporal transformer that reduces model size while increases the generalization of the model. The code and models will be made publicly available at \url{https://github.com/qtwang0035/IIP-Transformer}.

Overall, our contributions can be summarized as follows:
\begin{quote}
	\begin{itemize}
		\item We introduce the concept of body parts into transformer-based skeleton action recognition. Our proposed partition encoding significantly reduces self-attention computational complexity and is relatively insensitive to joint noise. 
		\item We propose IIP-Transformer, a novel spatial-temporal transformer network that captures intra-part and inter-part relations simultaneously.
		\item Extensive experiments on two large-scale skeleton action datasets, \eg NTU RGB+D 60 \& 120, show that our proposed IIP-Transformer achieves state-of-the-art performance with $ 2\sim36\times $ less computational cost.
	\end{itemize}
\end{quote}

\section{Related Work}

\noindent\textbf{Transformer.} In recent years, with the development of Natural Language Processing (NLP) tasks, the Transformer structure~\cite{vaswani2017attention} has been proposed to replace the traditional NLP network structures, \eg, RNNs. Different with  RNN architectures, the encoder and decoder of transformers completely rely on the self-attention mechanism, which can effectively solve the problems of long-sequence modeling and parallel processing. Because of these characteristics of the self-attention mechanism, the Transformer networks have been also applied in various computer vision tasks\cite{bello2019attention, dosovitskiy2020image,DBLP:conf/eccv/CarionMSUKZ20,plizzari2021skeleton} and achieve superior results comparing with the CNN models.\medskip

\noindent\textbf{Skeleton-based Action Recognition.} Skeleton-based action recognition has received increasing attentions due to its compactness comparing with the RGB-based representations. Previous data-driven methods rely on manual designs of traversal rules to transform the raw skeleton data into a meaningful form such as a point-sequence or a pseudo-image, so that they can be fed into the deep networks such as RNNs or CNNs for feature extraction\cite{lev2016rnn, wang2017modeling, cheron2015p, li2018co}. Inspired by the booming graph-based methods, Yan \etal~\cite{DBLP:conf/aaai/YanXL18} introduce GCN into the skeleton-based action recognition task, and propose the ST-GCN to model the spatial configurations and temporal dynamics of skeletons simultaneously. The GCN-based methods\cite{shi2019two, Cheng_2020_CVPR} use the topological structure of the human skeleton to aggregate features of related skeleton nodes and time series. Therefore, the GCN-based methods show better performance than previous methods. Instead of formulating the skeleton data into the images or graphs, Transformer-based methods directly model the dependencies of joints with pure attention blocks. Plizzari \etal~\cite{plizzari2021skeleton} propose a method that introduces Transformer in skeleton activity recognition and combine it with GCN. Shi \etal~\cite{shi2020decoupled} employ a solely transformer network to exploit relations between joints. Our proposed method utilizes part-level input, thus could efficiently capture both joint-level and part-level relations.\medskip

\begin{figure*}
	\centering
	\includegraphics[width=0.8\linewidth]{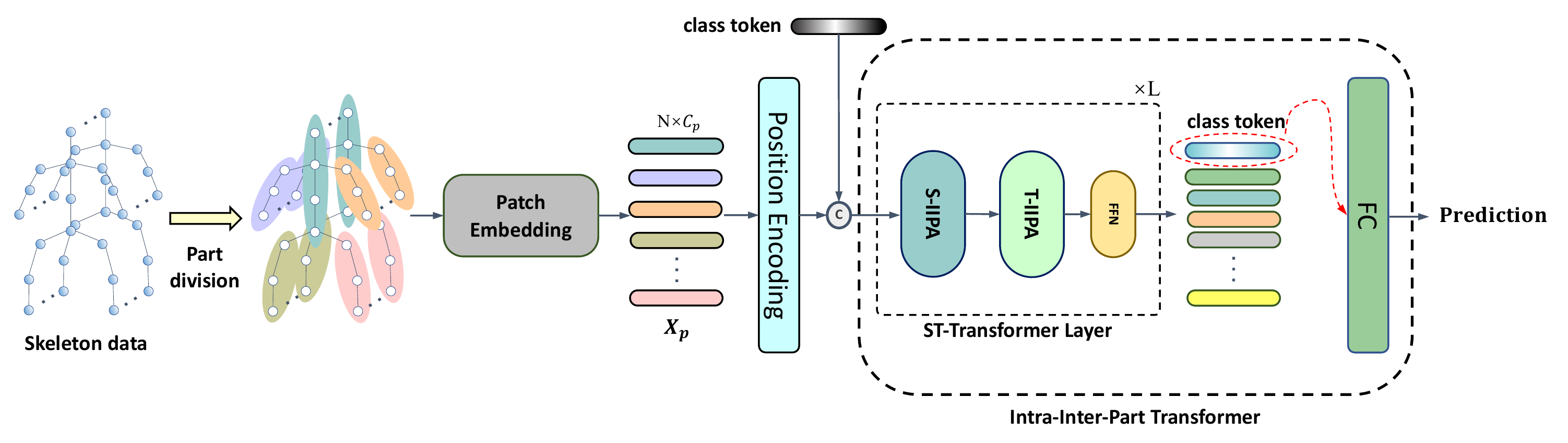} 
	\caption{The overall architecture of the proposed pipeline which is composed of Partition Encoding and IIP-Transformer.}
	\label{fig:2}
\end{figure*}

\noindent\textbf{Part-based Methods.} Part-based methods are designed to extract features of body parts individually since human body is a natural topology with five major parts. Thakkar \etal~\cite{thakkar2018part} divide graph into several sub-graphs with shared nodes. They employ GCN operation within sub-graphs of body parts and then propagate information between sub-graphs via the shared nodes. Huang \etal~\cite{huang2020part} propose an automatic partition strategy and utilize a part-based GCN to explore discriminative features from joints and body parts. Song \etal~\cite{song2020stronger} propose part-attention mechanism to discover the most informative parts. All these part-based methods employ complex strategies to propagate information individually or fuse information from all parts, while our work focuses on simultaneously capturing discriminative features from intra and inter parts with less computational cost.

\section{Methods}
\subsection{Overall Architecture}
\Cref{fig:2} shows the overall architecture of our pipeline, which mainly contains two sections, the partition encoding and the IIP-Transformer network. For an input skeleton sequence with $ V $ joints, $ F $ frames and $ C $ channels, we first divide $ V $ joints into $ P $ parts, and then a patch embedding layer is used to project these part data into $ N $ tokens with dimension $ C_p $, where $ N = P\times F $. Before being fed into IIP-Transformer, a class-token will be concatenated with tokens above, resulting in $ N + 1 $ tokens. There are $ L $ layers stacked in IIP-Transformer in total and each layer is composed of a spatial IIPA module (S-IIPA) that models the spatial relations between parts in the same frame, a temporal IIPA module (T-IIPA) that captures the temporal relations of parts among different frames and a feed-forward network. Each layer of IIP-Transformer maintains the same number of tokens, and thus we get $ N + 1 $ tokens as final output features. We feed the feature corresponding to the class-token into a fully-connected layer to obtain the classification scores. The implementation details will be introduced in the following sections.

\subsection{Partition Encoding}
\label{sec:3.2}

\begin{figure}[htb]
	\centering
	\includegraphics[width=0.6\linewidth]{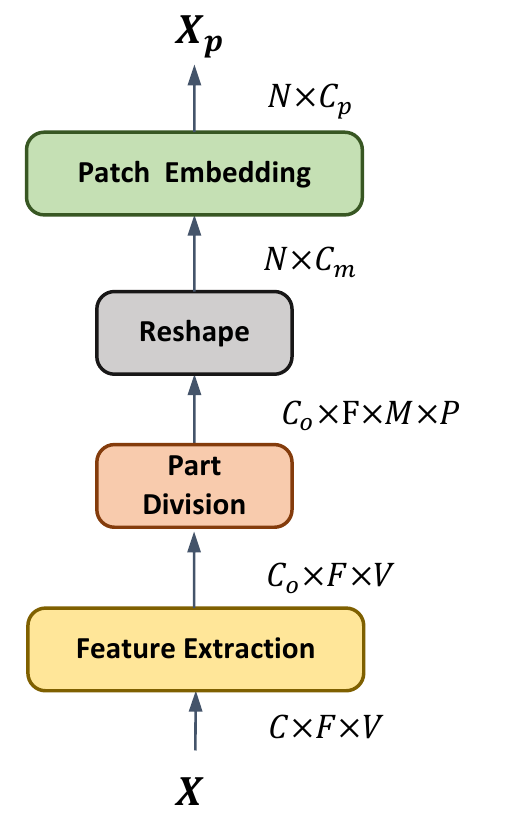}
	\caption{Illustration of the Partition Encoding Procedure.}
	\label{fig:3}
\end{figure}

In order to reduce the computation complexity of self-attention in spatial dimension, we propose a strategy to encode the joints into $P$ parts (5 parts in our case specifically). The partition encoding procedure is illustrated in \Cref{fig:3}. First, the raw skeleton sequence $X\in R^{ C\times F\times V} $ is fed into feature extraction layers $ f_J(\cdot) $ to obtain a deeper feature $X_J\in R^{C_o\times F\times V}$, where $C_o $ denotes the channel of features after feature extraction and $ f_J(\cdot) $ is implemented by two convolution layers with BatchNorm and ReLU function. Then $P$ individual body parts are obtained by selecting corresponding joints:
\begin{equation}
	X_J \rightarrow [x_1,x_2,\cdots,x_P], x_i\in R^{ C_o\times F\times M}
\end{equation}
where $M$ denotes the max number of joints in each part. The parts with less than $ M $ joints will be padded with zero. Subsequently, we concatenate $P$ parts and permute the dimensions:
\begin{equation}
	\widetilde{X_{J}} = Concat\left({x_i|i = 1,2,\cdots,P}\right)
\end{equation}
\begin{equation}
	\widetilde{X_{J}}\in R^{ C_o\times F\times M \times P}\rightarrow R^{N\times C_m} 
\end{equation}
where $N= P\times F $ and $C_m=C_o\times M$. Finally, $\widetilde{X_{J}}$ is fed into patch embedding layer $ f_P(\cdot) $ to aggregate information of the internal skeleton joints of a part and we get the final partition encoding $X_P\in R^{N \times C_P}$, where $C_P$ is the channels of the partition encoding and $ f_P(\cdot) $ is implemented by linear layers with ReLU function.

By aggregating the information of the internal skeleton joints of the parts, an informative encoding is extracted. It drives the model to concentrate on body parts instead of joints, and thus reduce the influence of individual noisy joints. Besides, experiments show that the partition encoding enables model to take sparser temporal inputs. 

\subsection{Intra-Inter-Part Transformer}

The backbone of IIP-Transformer is constructed by stacking $L$ layers of Spatial-Temporal Transformer Layer which is composed of S-IIPA, T-IIPA and feed-forward network. The core of S-IIPA and T-IIPA is Intra-Inter-Part Self-Attention mechanism. In this section, the IIPA mechanism and Spatial-Temporal Transformer will be briefly introduced.\medskip

\begin{figure}[htb]
	\centering
	\includegraphics[width=0.8\linewidth]{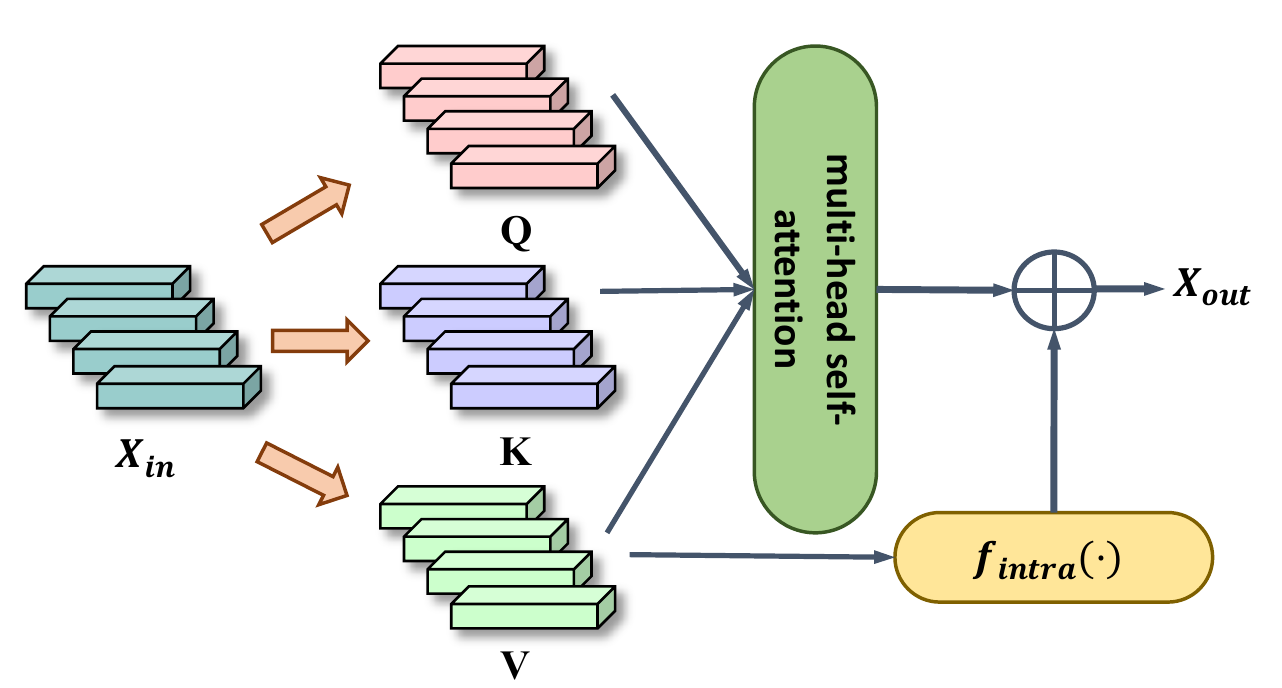}
	\caption{Illustration of the IIPA mechanism. The original multi-head self attention is used to explore relations between tokens while the function $ f_{intra} $ is used to capture internal relations of a token.}
	\label{fig:4}
\end{figure}

\noindent\textbf{Intra-Inter-Part Self-Attention.} Basic transformer utilizes self-attention mechanism to model the relationships between input tokens. But in part-level cases, each token represents a collection of joints in the same part. Therefore, the relationships of joints in the same part have not been fully exploited and can not be propagated effectively. We present a new self-attention mechanism named Intra-Inter-Part self-attention (IIPA) to simultaneously capture relationships inside and between tokens. As shown in \Cref{fig:4}, our proposed IIPA mainly consists of a standard multi-head self-attention and an intra-part branch.

Given the input feature $ X_{in}$, we first compute the query $Q$, key $K$ and value $V$ using three linear projection layers. The information flow across tokens is achieved by the standard multi-head self-attention:
\begin{equation} 
	X_{inter} = MHSA\left(Q, K, V\right)
\end{equation}
The intra-part branch $f_{intra}(\cdot)$ is implemented by linear layers. It takes value $V$ as input and extracts features inside tokens, termed joint-level feature.
\begin{equation}
	X_{intra} = f_{intra}\left(V\right)
\end{equation}
Finally, we fuse the multi-head self-attention output $X_{inter}$ with the intra-part branch output $X_{intra}$, so that features extracted by IIPA carry both joint-level and part-level information.
\begin{equation}
	X_{out} = X_{inter} + X_{intra}
\end{equation}

Comparing with standard self-attention mechanism, our proposed IIPA introduces no more than 1/4 calculations, but achieves obvious improvement in fine-grained actions, as shown in ablation study.\medskip

\begin{figure}[htb]
	\centering
	\includegraphics[width=0.8\linewidth]{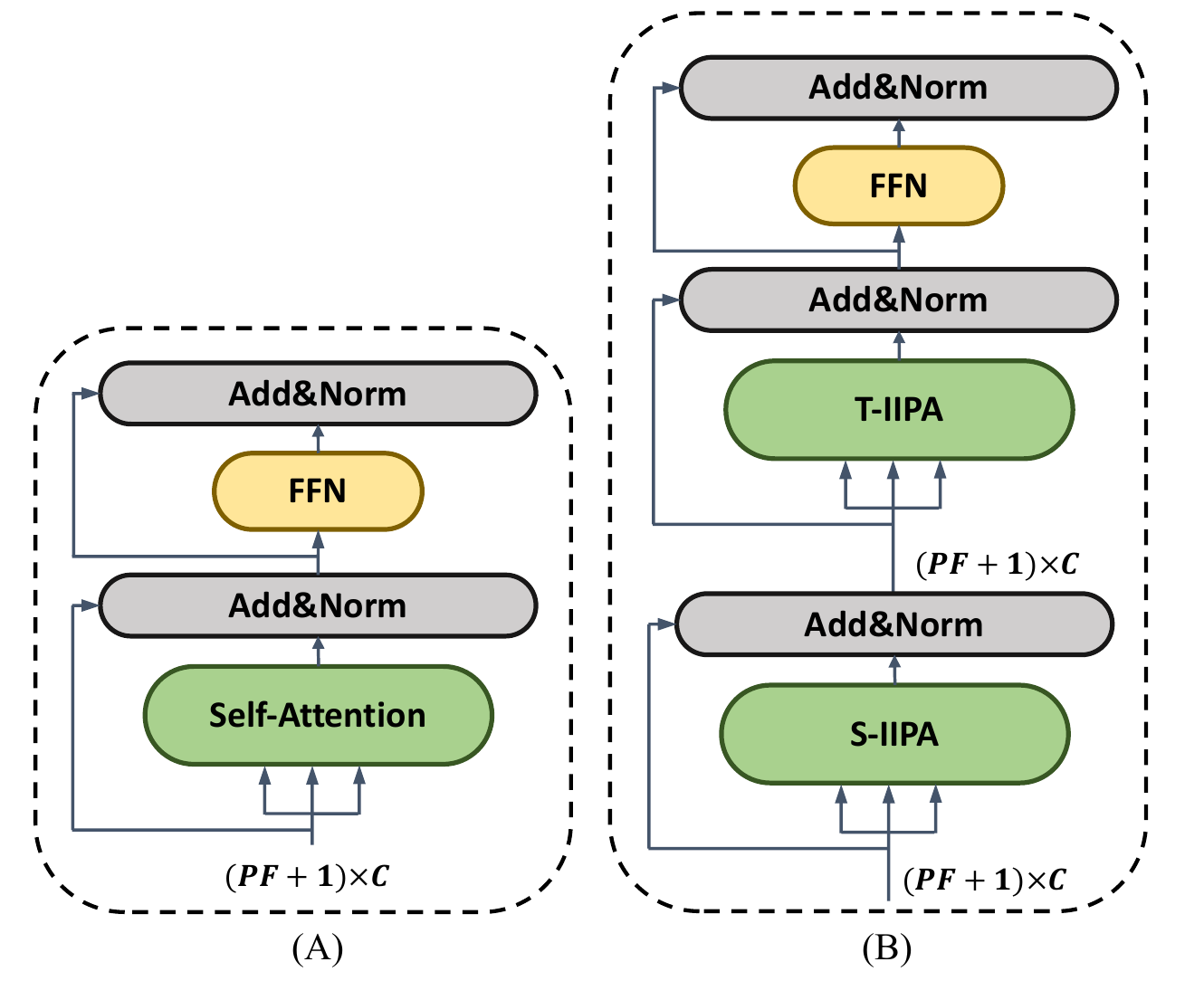}
	\caption{Two Transformer structures. (A) Flatten the spatial-temporal data into a single sequence and use an standard transformer. (B) Explore spatial and temporal relations with S-IIPA and T-IIPA respectively and fuse them by a feed-forward layer.}
	\label{fig:5}
\end{figure}

\noindent\textbf{Spatial-Temporal Transformer Layer.} In previous skeleton-based action recognition transformers\cite{shi2019skeleton,plizzari2021skeleton}, all features extracted by backbone is average-pooled to obtain the final feature for classification. Inspired by BERT~\cite{devlin2018bert}, class-token is introduced into our model. In BERT, the input $X_{1d}\in R^{(N+1)\times C}$ is a single dimension sequence with $N$ tokens and a class-token. The single dimension structure naturally fits the self-attention mechanism. But for the skeleton sequence which has two dimensions, there are two problems to solve: one is how to exploit spatial and temporal dimension with self-attention mechanism, another is how to deal with class-token. The first method is flattening the raw skeleton sequence $X_{2d} \in R^{P\times F\times C}$ into a single dimension sequence $X_{flatten} \in R^{PF \times C}$ and concatenating it with a class-token to obtain $X_{in}\in R^{(PF+1)\times C}$, where $P$ denotes the number of parts and $F$ denotes the number of frames. The output $X_{out}\in R^{\left(PF+1\right)\times C}$ is calculated by a standard transformer and the corresponding structure is shown in \Cref{fig:5}-(A):
\begin{equation}
     X_{out}=softmax\left(\frac{QK^T}{\sqrt C}\right)V
\end{equation}

However, it is unreasonable to treat spatial and temporal features equivalently since the semantic information they contain are totally different. Besides, the computational complexity of calculating the attention map is quadratic to sequence length, therefore flattening the skeleton sequence into a long single dimension sequence will introduce a large amount of calculations. 

\begin{figure}
	\centering
	\includegraphics[width=0.8\linewidth]{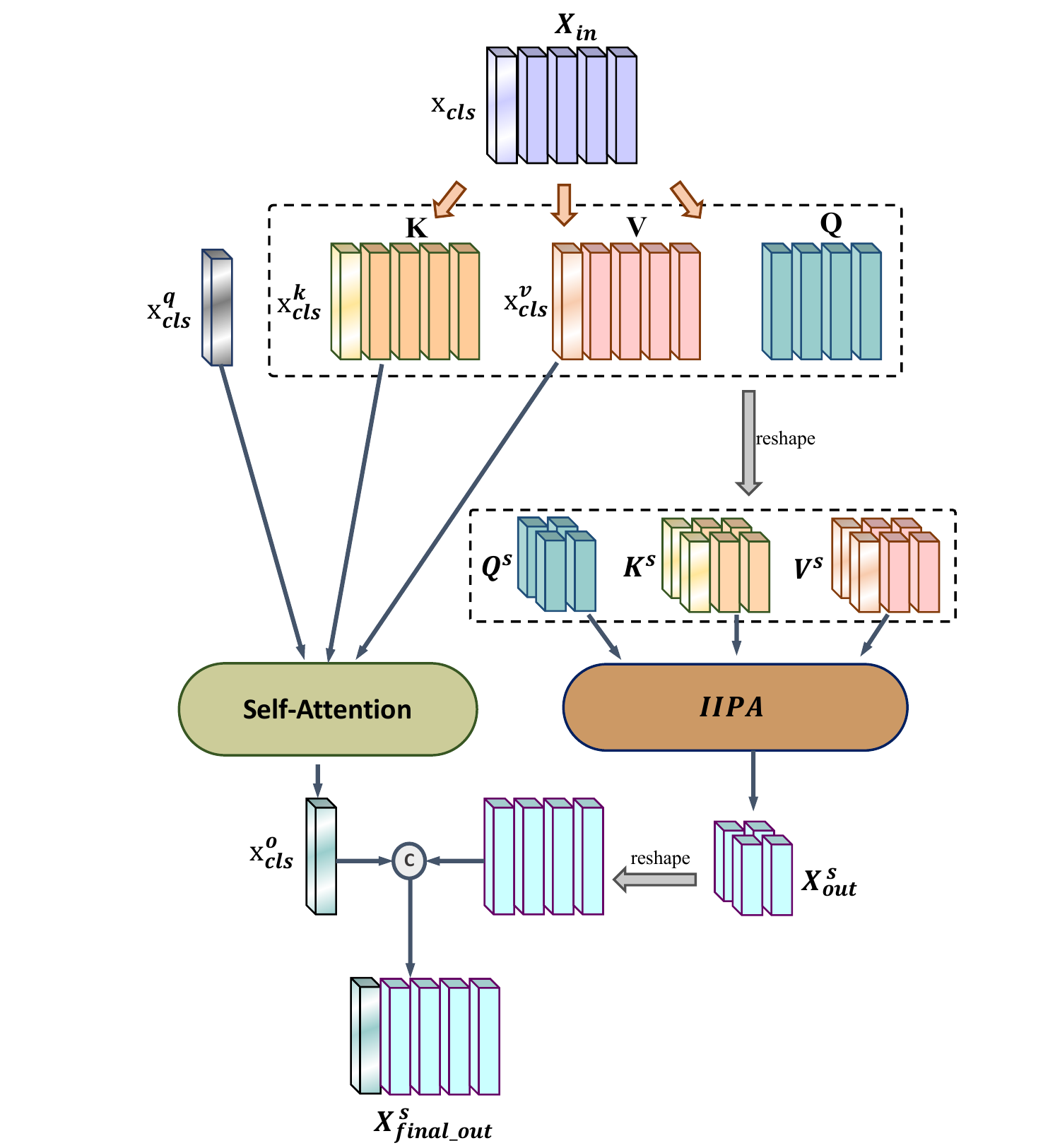}
	\caption{Illustration of S-IIPA. The symbol C stands for concatenate operation. T-IIPA has the same structure as S-IIPA except for reshape size of $ K,Q,V $.}
	\label{fig:6}
\end{figure}

Instead, the spatial and temporal dimension are treated as two different dimensions in Spatial-Temporal Transformer layer and processed by S-IIPA and T-IIPA respectively, as shown in \Cref{fig:5}-(B). The input of S-IIPA and T-IIPA is the same as above, a flattened sequence with a class-token. S-IIPA is adopted to explore the dependencies of parts in a single frame while T-IIPA is employed to propagate information for each part across different frames. Let's take S-IIPA for an example, we first project the part-level sequence $X_{in}\in R^{(PF+1)\times C}$ with three linear projections layers $f_q(\cdot),f_k(\cdot)$ and $f_v(\cdot)$:
\begin{equation}
    Q_{cls},Q=f_q\left(X_{in}\right)
\end{equation}
\begin{equation}
    K_{cls},K=f_k\left(X_{in}\right)
\end{equation}
\begin{equation}
    V_{cls},V=f_v\left(X_{in}\right)
\end{equation}
where $ Q_{cls},K_{cls},V_{cls} \in R^{1\times C} $ and $ Q, K, V \in R^{PF\times C} $. Due to the existence of class-token, we employ two branches, the self-attention branch and IIPA branch, to model class-token and other tokens separately, as shown in \Cref{fig:6}. The self-attention branch is similar to the above method, and thus the output $X_{cls}^o\in R^{1\times C}$ could obtain information from all parts at all frames.
\begin{equation}
    K_c = Concat\left(K_{cls}, K\right)
\end{equation}
\begin{equation}
    V_c = Concat\left(V_{cls}, V\right)
\end{equation}
\begin{equation}
    X_{cls}^o = softmax\left(\frac{Q_{cls}\left(K_{c}\right)^T}{\sqrt C}\right)V_{c}
\end{equation}
In the IIPA branch, we reshape $K,Q,V$ from single dimension into temporal$\times$spatial dimension and repeat $K_{cls}, Q_{cls}, V_{cls}$ in the temporal dimension, then concatenate them in the spatial dimension.
\begin{equation}
    Q,K,V\in R^{PF\times C}\rightarrow Q^s,K^s,V^s\in R^{F\times P\times C}
\end{equation}
\begin{equation}
    K_{cls},V_{cls} \in R^{1\times C}\rightarrow R^{F\times1\times C}
\end{equation}
\begin{equation}
    K^s=Concat\left(K_{cls},\ K^s\right)
\end{equation}
\begin{equation}
    V^s=Concat\left(V_{cls},\ V^s\right)
\end{equation}
By following the preceding steps, we get $K^s,V^s\in R^{F\times\left(P+1\right)\times C}$ , ${\left(P+1\right)}$ means appending a class-token in every frame. Then $ Q^s, K^s, V^s $ are fed into IIPA to calculate the output frame by frame, and each frame uses a unique attention map:
\begin{equation}
    X_f=softmax\left(\frac{Q_f^s\left({K_f^s}\right)^T}{\sqrt C}\right)V_f^s,\ f=1,2\cdots,F
\end{equation}
where $X_f\in R^{P\times C}$ is the output for frame $f$, $Q_f^s\in R^{P\times C};\ K_f^s,\ V_f^s\in R^{\left(P+1\right)\times C}$. Finally, we concatenate the outputs of all frames and reshape it back to single dimension:
\begin{equation}
    X_{out}^s=Concat\left(\left\{X_f|f=1,2,\cdots,F\right\}\right)
\end{equation}
\begin{equation}
    X_{out}^s\in R^{F\times P\times C}\rightarrow R^{FP\times C}
\end{equation}
\begin{equation}
    X_{final\_out}^s=Concat\left(X_{cls}^o,X_{out}^s\right)
\end{equation}
where $X_{final\_out}^s\in R^{\left(FP+1\right)\times C}$ is the final output of S-IIPA with the same dimension size as the input sequence. The only difference in T-IIPA is that $Q^t\in R^{P\times F\times C};K^t,V^t\in R^{P\times\left(F+1\right)\times C}$, corresponding to $Q^s,K^s,V^s$, thus the output is calculated part by part.

Comparing with other spatial-temporal transformer methods, \eg DSTA~\cite{shi2020decoupled}, which uses two complete transformers to model space and time, we remove the feed-forward layer in the spatial transformer structure and combine S-IIPA, T-IIPA and feed-forward layer as a new spatial-temporal transformer structure. This structure performs spatial and temporal feature extraction in one stage so that the same-order features can be fully exploited. Meanwhile, it also reduces a large amount of parameters in feed-forward layer and improves the generalizability of the model.

\subsection{Data Augmentation}
Rao \etal~\cite{rao2021augmented} propose several joint-level skeleton data augmentation methods, such as \textit{Rotation}, \textit{GaussianNoise}, \textit{GaussianBlur}, \textit{JointMask}, to improve the generalizability of models. However, since the part-level encoding is more robust to joint noises, these methods do not work well on part-level skeleton data, except for \textit{Rotation}. Therefore we propose a new data augmentation method called \textit{PartMask}, to encourage the model to reason globally instead of relying on a particular part. The \textit{Rotation} and \textit{PartMask} methods are defined as follows.\medskip

\noindent\textbf{Rotation.} Most public skeleton-based datasets are captured in specific view points. We obtain plentiful data through \textit{Rotation} transformation which simulates viewpoint change of the camera.

Based on Euler’s rotation theorem, any 3D rotation can be disassembled into a composition of rotations about three axes~\cite{wang2017modeling}. The basic rotation matrices are represented as below:

\begin{equation}
    R_{x}(\alpha)=\left[\begin{array}{ccc}
1 & 0 & 0 \\
0 & \cos \alpha & -\sin \alpha \\
0 & \sin \alpha & \cos \alpha
\end{array}\right]
\end{equation}
\begin{equation}
    R_{y}(\beta)=\left[\begin{array}{ccc}
\cos \beta & 0 & \sin \beta \\
0 & 1 & 0 \\
-\sin \beta & 0 & \cos \beta
\end{array}\right]
\end{equation}
\begin{equation}
    R_{z}(\gamma)=\left[\begin{array}{ccc}
\cos \gamma & -\sin \gamma & 0 \\
\sin \gamma & \cos \gamma & 0 \\
0 & 0 & 1
\end{array}\right]
\end{equation}
\begin{equation}
    R=R_{z}(\gamma) R_{y}(\beta) R_{x}(\alpha)
\end{equation}
where $ R_{x}(\alpha),R_{y}(\beta),R_{Z}(\gamma) $ denote the rotation matrices of $ x,y,z $ axis with angle $ \alpha,\beta,\gamma $ respectively, and $ R $ is the general rotation matrix that will be applied to original coordinates of the skeleton sequence. In this work, the rotation angles are randomly sampled from $ [-\pi/10, \pi/10] $.\medskip

\noindent\textbf{PartMask.} As an effective augmentation strategy to reduce reliance on specific regions, the mask strategy has been widely used in data augmentation. But simply employing a random zero-mask to a number of body joints in skeleton frames before partition encoding, which is similar to joint-level noises, does not work well due to the anti-noise ability of partition encoding. On the other hand, capturing the global information instead of focusing on a particular part will benefit the action classification tasks. Therefore we employ a part-level mask to encourage the model to reason globally. Specifically, we randomly select a certain body part $ p $ from $ [1,2,\cdots,P] $, and apply zero-mask to it in all frames:
\begin{equation}
\begin{split}
X_J &= X_J \odot mask \\
    &= [x_1,x_2,\cdots,x_P] \odot [1,1,\cdots,\mathop{0}_{\overline{p}},\cdots,1] \\
\end{split}
\end{equation}
where $ X_J $ denotes the intermediate result of partitioning, as discussed in \Cref{sec:3.2}, and mask is an all-ones vector except for the position of the selected part $ p $.

\section{Experiments}
\subsection{DataSets}
\noindent\textbf{NTU RGB+D.} NTU RGB+D~\cite{Shahroudy_2016_CVPR} is a widely used large-scale human skeleton-based action recognition dataset, which contains 56,880 skeletal action sequences. These action sequences were performed by 40 volunteers and divided into 60 categories. Each action sequence is completed by one or two subjects and is captured by three Microsoft Kinect-V2 cameras from different views simultaneously. The benchmark evaluations include Cross-Subject (X-Sub) and Cross-View (X-View). In the Cross-Subject, training data comes from 20 subjects, and testing data comes from the other 20 subjects. In the Cross-View, training data comes from camera views 2 and 3, and testing data comes from camera view 1. Note that there are 302 wrong samples that need to be ignored.\medskip

\noindent\textbf{NTU RGB+D 120.} NTU RGB+D 120~\cite{8713892} is currently the largest human skeleton-based action recognition dataset. It is an extension of the NTU RGB+D dataset, with 113,945 action sequences and 120 action classes in total. These action sequences were performed by 106 volunteers, captured with three cameras views, and contains 32 setups, each of which represents a different location and background. The benchmark evaluations include Cross-Subject (X-Sub) and Cross-Setup (X-Setup). In the Cross-Subject, training data comes from 53 subjects, and testing data comes from the other 53 subjects. In the Cross-Setup, training data comes from samples with even setup IDs, and testing data comes from samples with odd setup IDs. In this dataset, 532 bad samples should be ignored.

\subsection{Implementation Details}
All experiments are conducted on 8 GTX 1080Ti GPUs. Our model is trained using SGD optimizer with momentum 0.9 and weight decay 0.0002. The training epoch is set to 300. Learning rate is set to 0.1 and decays with a cosine scheduler. The batch size is 64 and each sample contains 32 frames. The number of Spatial-Temporal Transformer layer is set to 4.

\subsection{Ablation Study}
In this section, we investigate the effectiveness of the proposed components of the IIP-Transformer. All experiments are conducted on NTU RGB+D 60 with joint stream if no special instruction.\medskip

\begin{figure}[t]
	\centering
	\includegraphics[width=0.8\linewidth]{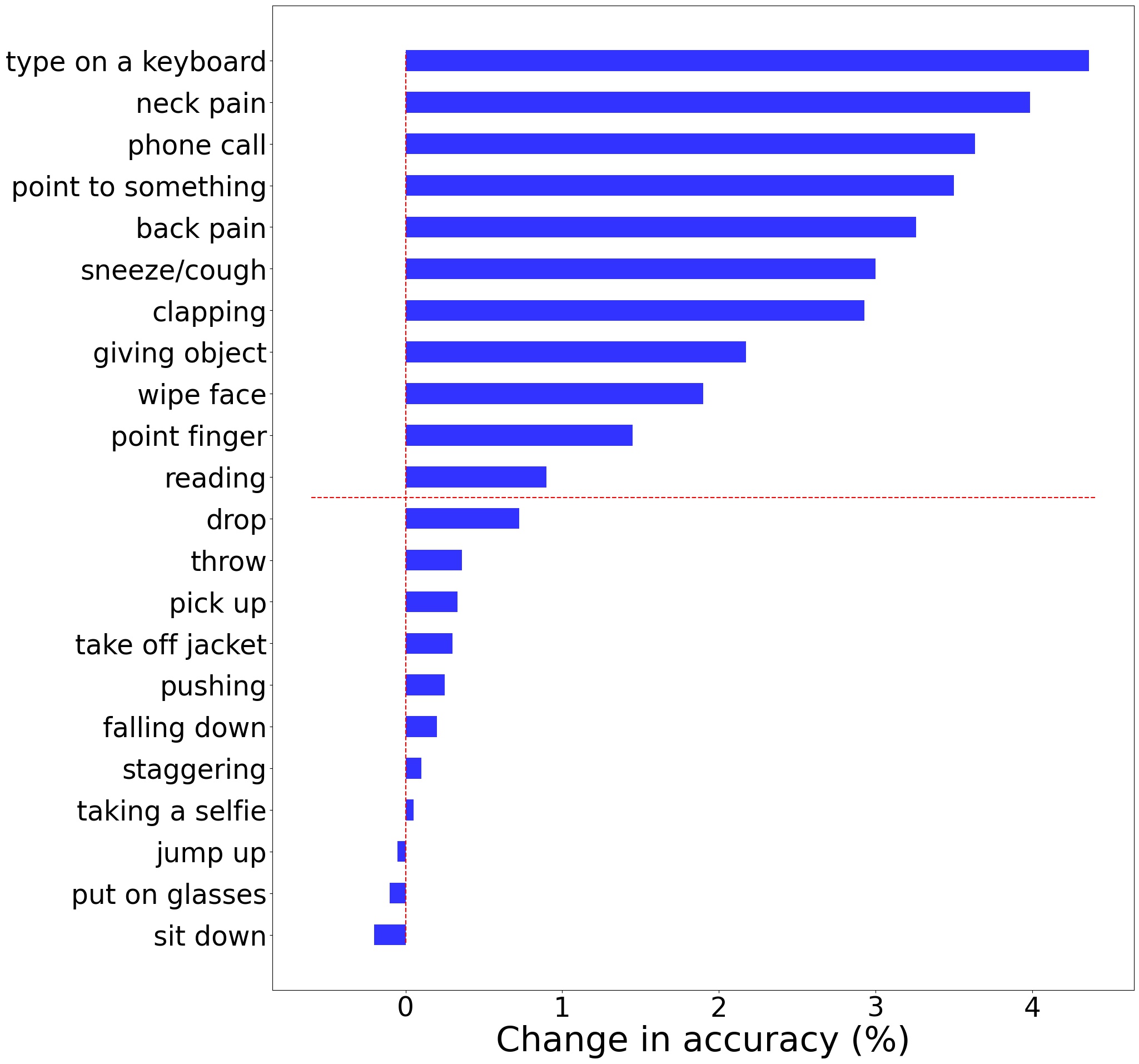}
	\caption{The accuracy comparison on actions with different motion intensity. The categories above the red horizontal line are fine-grained action categories.}
	\label{fig:7}
\end{figure}

\noindent\textbf{Effect of IIPA.} The comparison between IIPA and standard Self-Attention in \Cref{tab:1} shows that there are 1.8\% and 2\% improvements in X-Sub and X-View respectively. We pick out some actions and visualize the changes of accuracy between IIPA and standard Self-Attention in Figure \Cref{fig:7}. It demonstrates that the IIPA achieves remarkable improvements on fine-grained actions (e.g., type on a keyboard, clapping, reading etc.), while the accuracy of drastic actions (e.g., throw, pushing, pickup etc.) is about the same, which indicates that our proposed IIPA can exploit the joint-level information of body parts more effectively.\medskip

\begin{table}
  \centering
  \begin{tabular}{l|cc}
    \hline \textbf{Methods} & \textbf{X-Sub} & \textbf{X-View}\\
    \hline
    \hline Standard Self-Attention & 85.3 & 91.2 \\
    \hline IIPA (ours) & \textbf{87.1} & \textbf{93.2} \\
    \hline
  \end{tabular}
  \caption{Comparison between IIPA and standard Self-Attention.}
  \label{tab:1}
\end{table}

\noindent\textbf{Effect of Class-Token.} To explore the effect of class-token, we replace it by simply performing global average pooling to the output features from last layer\cite{shi2020decoupled, plizzari2021skeleton}. As shown in \Cref{tab:2}, by introducing the class-token, the performance of X-Sub and X-View improve by 1.3\% and 1.7\% respectively.

\begin{table}[h]
	\centering
	\begin{tabular}{l|cc}
		\hline \textbf{Methods} & \textbf{X-Sub} & \textbf{X-View} \\
		\hline
		\hline Non-CLS & 85.8 & 91.5 \\
		\hline CLS (ours) & \textbf{87.1} & \textbf{93.2} \\
		\hline
	\end{tabular}
	\caption{Ablation study on class-token. Non-CLS denotes the previous global average pooling method.}
	\label{tab:2}
\end{table}

\begin{figure}[t]
	\centering
	\includegraphics[width=0.8\linewidth]{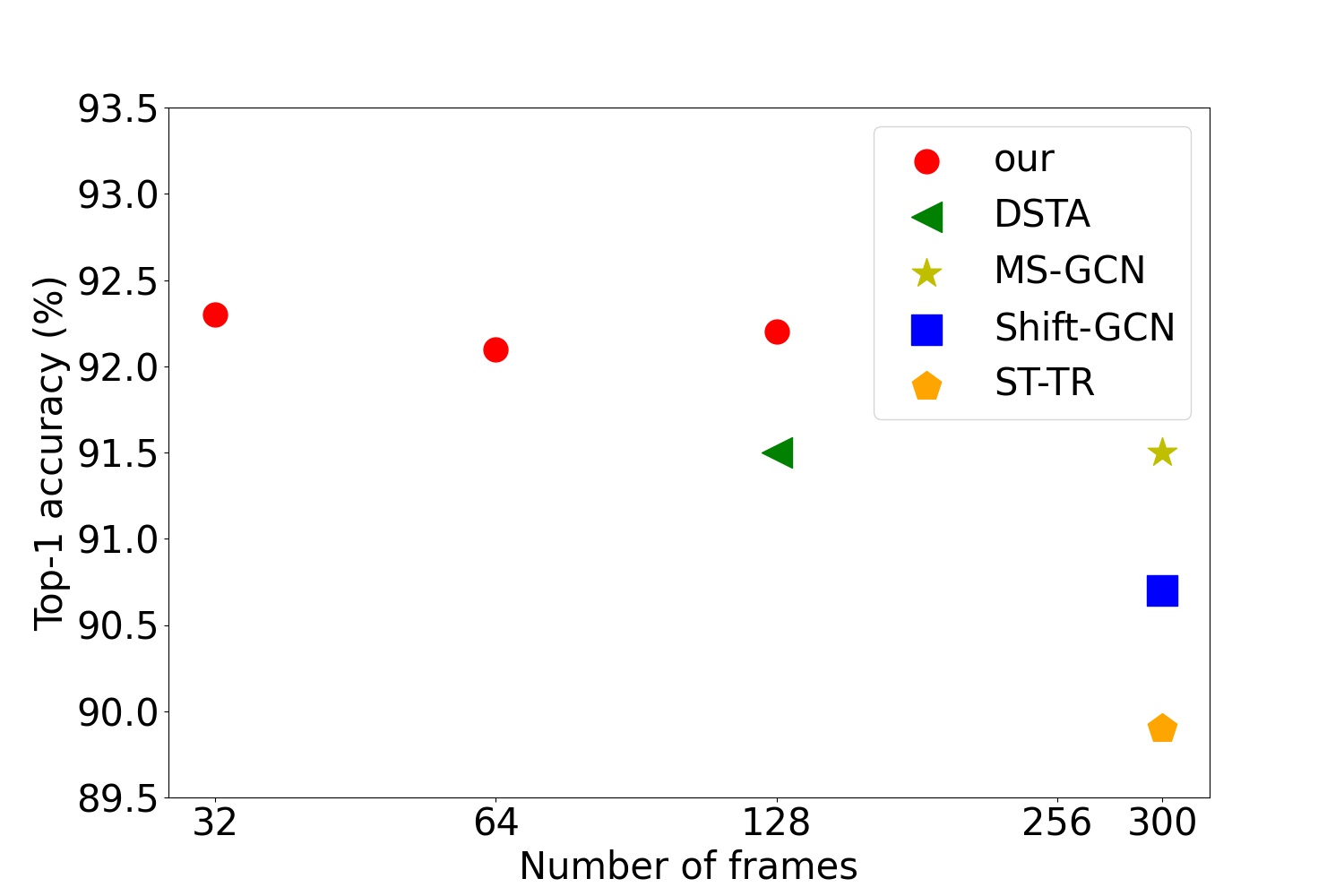}
	\caption{Ablation study over different methods with different input frames on NTU RGB+D 60 (X-Sub). The results are obtained after multi-stream fusion which will be introduced in \Cref{sec:4.4}.}
	\label{fig:8}
\end{figure}

\noindent\textbf{Effect of Partition Encoding.} To explore the effect of Partition Encoding, we remove the Partition Encoding module from the proposed pipeline, which increases the number of tokens by 5 times (from 5 parts to 25 joints) and conduct experiments with different frame numbers. Comparing with Non-Partition Encoding (Non-PE), Partition Encoding (PE) improves accuracy by 1.1\% with much less computational cost, as shown in \Cref{tab:3}. Besides, PE enables the model to take sparser frames as temporal inputs, while the accuracy of Non-PE drops by 1.9\% when reducing the number of frames from 128 to 32, as shown in \Cref{tab:3}. In addition, we compare the accuracy of methods\cite{Cheng_2020_CVPR,liu2020disentangling,shi2020decoupled,plizzari2021skeleton} with different number of input frames, as shown in \Cref{fig:8}. Our IIP-Transformer achieves comparable results with only 32 frames.\medskip

\begin{table}
	\centering
	\begin{tabular}{l|ccc}
    \hline \textbf{Methods} & \textbf{Frames}  & \textbf{X-Sub}  &  \textbf{FLOPs} \\
    \hline 
      &32&  84.1  &44.6G\\
     Non-PE &64&  85.2  &97.2G\\
      &128&  86.0  &198.7G\\
    \hline
      &32& \textbf{87.1}   & \textbf{7.2G}\\
     PE &64& 86.9   & 19.8G\\
      &128& 87.0   & 45.5G\\
    \hline
    \end{tabular}
    \caption{Ablation study of the Partition Encoding.}
    \label{tab:3}
\end{table}
\begin{table}
	\centering
	\begin{tabular}{l|cc}
		\hline
		\textbf{Methods} & \textbf{X-Sub} & \textbf{X-View} \\
		\hline
		\hline
		Origin		   	        &87.1  &93.2\\
		Gaussian-Noise	       	&87.1  &93.1\\
		Gaussian-Blur	       	&86.5  &92.8\\
		Joint-Mask	       	    &87.2  &92.9\\
		\hline
		Rotation                &87.9  &93.9\\
		Part-Mask               &88.4  &94.0\\
		\hline
		Rotation+Part-Mask  &\textbf{88.9} & \textbf{94.2}\\
		\hline
	\end{tabular}
	\caption{Ablation study of data augmentation on NTU RGB+D 60.}
	\label{tab:4}
\end{table}
\begin{table*}
    \centering
	\begin{tabular}{l|cc|cc|cc}
		\hline
		\textbf{Methods} &\textbf{X-Sub} &\textbf{X-View} &\textbf{X-Sub 120} &\textbf{X-Set 120} &\textbf{Param} &\textbf{FLOPs}\\
		\hline
		\hline
		ST-LSTM\cite{liu2016spatio} 		    	&69.2&77.7&55.7 &57.9 &-&-\\				
		HCN\cite{li2018co} 	        				&86.5 &91.1&-&- &-&-\\				
		\hline				
		ST-GCN\cite{DBLP:conf/aaai/YanXL18}		        	&81.5 &88.3&-&- &3.1M&16.3G\\
		2s-AGCN\cite{shi2019two}		      		&88.5 &95.1 &82.9 &84.9 &6.9M&37.3G\\
		AGC-LSTM\cite{si2019attention}				&89.2 &95.0&-&- &22.9M&-\\
		PL-GCN\cite{huang2020part}					&89.2 &95.0&-&- &20.7M&-\\
		DGNN\cite{shi2019skeleton}					&89.9 &96.1&-&- &26.2M&-\\		
		Shift-GCN\cite{Cheng_2020_CVPR}         	&90.7 &96.5 &85.9 &87.6 &2.8M&10.0G\\		
		DC-GCN+ADG\cite{cheng2020decoupling}		&90.8 &96.6 &86.5 &88.1 &4.9M&25.7G\\
		PA-ResGCN-B19\cite{song2020stronger} 		&90.9 &96.0 &87.3 &88.3 &3.6M&18.5G\\
		Dynamic GCN\cite{DBLP:journals/corr/abs-2007-14690} 			&91.5 &96.0 &87.3 &88.6 &14.4M&-\\
		MS-G3D\cite{liu2020disentangling} 		  	&91.5 &96.2 &86.9 &88.4 &2.8M&48.8G\\
		MST-GCN\cite{chen2021multi}					&91.5&\textbf{96.6} &87.5&88.8 &12.0M&-\\
		EfficientGCN-B4\cite{song2021constructing}	&91.7 &95.7 &88.3 &89.1 &\textbf{2.0M}&15.2G\\
		\hline
		ST-TR\cite{plizzari2021skeleton}			&89.9 &96.1 &82.7 &84.7 &12.1M&259.4G\\
		DSTA\cite{shi2020decoupled}					&91.5 &96.4 &86.6 &89.0 &4.1M&64.7G\\
		\hline
		\hline
		\textbf{IIP-Transformer}	&\textbf{92.3} &96.4 &\textbf{88.4}&\textbf{89.7} &2.9M &\textbf{7.2G}\\
		\hline
	\end{tabular}
	\caption{Comparison of top-1 accuracy (\%), model size and computational complexity over different methods on the NTU RGB+D 60/120 datasets.}
	\label{tab:5}
\end{table*}

\noindent\textbf{Effect of Data Augmentation.} To evaluate the impact of the proposed data augmentation strategies, we conduct experiments with different data augmentation (see \Cref{tab:4}). Applying part-level data augmentation strategies(\eg, Rotation and Part-Mask) on X-Sub improves the accuracy by 0.8\% and 1.3\% respectively, and the best results are achieved when combining these two strategies. While the effect of joint-level data augmentation (\eg, Gaussian-Noise, Gaussian-Blur and Joint-Mask) is marginal. Intuitively, the reason that Part-Mask strategy can effectively improve the accuracy of the model is that it encourages the model to reason globally and reduce the dependency on any particular part.

\subsection{Comparison with State of the Arts}
\label{sec:4.4}
Similar to most SOTA methods, we follow the same multi-stream fusion strategies proposed in ~\cite{Cheng_2020_CVPR} for fair comparison. We train four models with different modalities, \eg, joint, bone, joint motion, and bone motion respectively, then average the \textit{softmax} outputs from multiple streams to obtain the final scores during inference. The comparison results are shown in \Cref{tab:5}.\medskip

\noindent\textbf{Accuracy Comparison.} First we compare our results with GCN-based methods\cite{DBLP:conf/aaai/YanXL18, shi2019two, shi2019skeleton, Cheng_2020_CVPR, cheng2020decoupling, song2020stronger, DBLP:journals/corr/abs-2007-14690, liu2020disentangling, song2021constructing}. Our proposed method outperforms the best GCN method by 0.6\% on X-sub of NTU RGB+D 60 and 0.6\% on X-Set of NTU RGB+D 120 respectively. In terms of Transformer-based methods\cite{plizzari2021skeleton, shi2020decoupled}, our method also surpasses them on most of the metrics by a significant margin.\medskip

\noindent\textbf{Complexity Comparison.} We evaluate the computational complexity with FLOPs, \eg the number of floating-point multiplication-adds, and measure the model size with the amount of parameters. The computational complexity of our proposed IIP-Transformer is 2.2 times less than ST-GCN~\cite{DBLP:conf/aaai/YanXL18} and 6.7 times less than MS-G3D~\cite{liu2020disentangling}, while the model sizes are similar. Comparing with the Transformer-based methods, IIP-Transformer achieves superior results with only a fraction of computational complexity and model size.

\section{Conclusion}
In this work, we propose a novel intra-inter-part transformer network (IIP-Transformer) for skeleton-based action recognition. It effectively captures inter-part and intra-part dependencies. Thanks to the part-level encoding and spatial-temporal separation, our method enjoys high efficiency. Besides, a part-level data augmentation named \textit{PartMask} is proposed to encourage the model focus on global parts. On two large scale datasets, NTU RGB+D 60 \& 120, the proposed IIP-Transformer notably exceeds the current state-of-the-art methods with $ 2\sim36\times $ less computational cost.

%%%%%%%%% REFERENCES
{\small
\bibliographystyle{ieee_fullname}
\bibliography{egbib}
}

\end{document}